# Xu: An Automated Query Expansion and Optimization Tool

Morgan Gallant
School of Computing
Queen's University
Kingston, ON, Canada
mgallant@cs.queensu.ca

Haruna Isah
School of Computing
Queen's University
Kingston, ON, Canada
isah@cs.queensu.ca

Farhana Zulkernine
School of Computing
Queen's University
Kingston, ON, Canada
farhana@cs.queensu.ca

Shahzad Khan
Gnowit Inc.
Ottawa, ON, Canada
shahzad@gnowit.com

*Abstract*— The exponential growth of information on the Internet is a big challenge for information retrieval systems towards generating relevant results. Novel approaches are required to reformat or expand user queries to generate a satisfactory response and increase recall and precision. Query expansion (QE) is a technique to broaden users' queries by introducing additional tokens or phrases based on some semantic similarity metrics. The tradeoff is the added computational complexity to find semantically similar words and a possible increase in noise in information retrieval. Despite several research efforts on this topic, QE has not yet been explored enough and more work is needed on similarity matching and composition of query terms with an objective to retrieve a small set of most appropriate responses. QE should be scalable, fast, and robust in handling complex queries with a good response time and noise ceiling. In this paper, we propose Xu, an automated QE technique, using high dimensional clustering of word vectors and Datamuse API, an open source query engine to find semantically similar words. We implemented Xu as a command line tool and evaluated its performances using datasets containing news articles and human-generated QEs. The evaluation results show that Xu was better than Datamuse by achieving about 88% accuracy with reference to the human-generated QE.

*Keywords-boolean query; datamuse api; high-dimensional clustering; information retrieval; search engine; user query*

## I. INTRODUCTION

Efficient search for information from distributed knowledge management systems or the World Wide Web (WWW) is a crucial part for corporations, governments, and general users. Information Retrieval (IR) as a research area focuses on the structure, analysis, organization, storage, searching, and retrieval of desired information from a collection of information sources [1, 2]. Queries are the primary ways in which the information seekers communicate with IR systems. In an ad-hoc IR setting, an information need from a user is expressed as a single word query or a more complex query expression delineated by Boolean operators. The query is then processed by IR applications to retrieve the best response [3]. A search engine is an IR system that accepts a query as input and returns a ranked list of results. Search techniques are used in many applications including content categorization [4], question answering, media monitoring, advertising, security, scientific discovery, intelligence analysis, decision support, and robotic systems.

A core issue in IR and search applications is the evaluation of search results. The emphasis is on users and their information needs. The users of an IR system such as a search engine, are the ultimate judges of the quality of the results [2]. There are many reasons why a search result may not meet user expectations. Sometimes the query expression may be too short to dictate what the user is looking for or may not be well formulated [5]. In most cases, the user's original query is not sufficient to retrieve the information that the user is looking for [6]. Spink et al. analyzed 1,025,910 search queries from 211,063 users and found out that the average number of terms per query is just 2.16 [7]. Brief or poorly constructed queries using informal languages can lead to semantic or lexical gaps and result in poor retrieval or multiple subsequent queries [8].

A simple query-response paradigm works well for simple searches, when the user has a full knowledge of the exact set of words to look for. However, the ability of this approach is limited as it only searches for exact matches. Query formulation is itself a search problem that involves searching through the space of possible queries rather than through the space of results [4, 5]. It is very difficult to define a set of words that embody everything that a user is looking for. Instead of striving to create an optimal ranking algorithm, developers of search engines should aspire to create an optimal query formulation and refinement mechanism [9]. This is where mechanisms for query optimization such as query suggestion and expansion can come into effect [2]. Query Expansion (QE), which is defined as the set of methodologies, algorithms or techniques for refining search queries, is often utilized for the purpose of retrieving more relevant results [10]. QE tries to improve search and IR by analyzing the relationships between the terms or words in a query, and other words in a lexical database such as Wordnet [11] in order to find potentially related words so that the original query is better represented [8]. QE is performed either by expanding the initial query through the addition of new related terms or by the selective retention of terms from the original query composed using effective Boolean logic expressions [5]. Code search [12] is a typical application area of QE. According to Lu et al., the words used in a query written by a software maintenance engineer are often different from the lexicon used by the developers [12]. To overcome this, the query needs to be reformulated. However, many existing code search techniques do not provide support for developers to rewrite a query, hence the need for further research on QE.

Several experimental studies on QE agree that the quality of the query results is enhanced by more than ten percent with the expansion of the user query [5]. While QE can improve the recall ratio of search results by attempting to retrieve many relevant documents, it can however, adversely affect precision. To achieve a better precision, Boolean operators such as AND, OR, NOT, WITH and NEAR are used to transform the expanded query into a Boolean query [4, 9]. A common issue with the frequently used Boolean operators (AND and OR) is that an AND operator improves precision but reduces recall rate whereas an OR operator reduces precision but improves recall rate [9]. QE should be scalable, fast, and robust in handling complex queries with a good response time and noise ceiling [6]. These are the interesting major challenges that this study aims to address.

### A. Contributions

The contributions of the paper are as follows. We present Xu, an automated QE technique, which we developed and implemented as a command line tool. To achieve query expansion and optimization, Xu utilizes (i) Datamuse [13], an open source Application Programming Interface (API) for finding similar or related words, (ii) a Wikipedia-trained Word2Vec model to generate word vectors for each word, (iii) a function for ranking and selecting top k suggestions from Datamuse by their vector similarity to the initial search query, (iv) a high dimensional clustering technique for grouping semantically similar vectors, and (v) a model for transforming the expanded queries to Boolean queries using Boolean logic operators. We validate our prototype tool for 373 unique user queries (out of which 241 include human-generated QEs) obtained from a media monitoring company. The evaluation results demonstrate that the expansions from Xu perform significantly better than those of Datamuse with an average accuracy of 88% as against the Datamuse QE approach which realized an accuracy of 70%.

### B. Use Case Scenario

Xu can be used as a part of a variety of different query and data processing systems and pipelines, the most common of which is a basic search engine for document retrieval from a data lake. A user query can be fed into Xu, and the expanded query can be used to search a corpus of web pages and documents present in the repository at the backend of the search engine. Any matching page(s), which meet all the filter criteria, would be returned to the user. Some adjustments, however, are required. Xu is designed to handle queries in the form of 'word and/or word', where 'word' can be a simple word such as 'paper' or a complex word such as 'academic paper'.

The rest of the paper is organized as follows. Section II presents a background on relevant concepts and the related work on QE and high-dimensional clustering. Section III presents an overview of the computational models used to build Xu, our automated QE and optimization tool, while the implementation is illustrated in section IV. The evaluation of Xu is presented in Section V. Section VI concludes the paper with a discussion of future work.

## II. BACKGROUND

Search engines support two major functions, indexing and query processing. The indexing process builds the structures to enable searching and data acquisition, performs data transformation as necessary, and creates the index. The query process utilizes these structures and executes users' queries to produce a ranked list of results. It involves user interaction, ranking, and evaluation [2]. This study focuses on the query processing aspect which is largely unexplored compared to the data indexing aspect. The performance of ranking in terms of efficiency depends on the indexing, while its effectiveness depends on the query processing and retrieval model. The most critical challenge in achieving the effectiveness is the issue of mismatch, a situation where indexers and users do not use the same words [5]. To deal with this problem, several approaches have been proposed including query transformation and refinement, relevance feedback, and clustering of the search results [6]. The next sub-sections describe these fundamental concepts of query processing.

### A. Query Transformation

Query transformation refers to the use of various techniques to improve the initial user query. Transformations that are usually carried out on text queries before ranking include tokenization, stop word removal, stemming, spelling check, query suggestion, and query expansion. Relevance feedback is one of the transformations that are usually carried out on text queries after ranking [2].

*1) Query Expansion*

In 1993, Qiu and Frei developed a probabilistic QE model [14] based on an automatically generated similarity thesaurus, defined as a matrix that consists of term-term similarities. Their study developed two key ideas, generating a model of synonyms/similar words, and using the 'most similar' of these words to append to the initial search query to increase accuracy. Qiu and Frei validated their ideas with query expansion using word lexical-semantic relations [15]. Voorhees utilized the WordNet system to generate similar words using various levels of query expansions. The results of the experiments revealed that aggressively expanded queries performed worse than the unexpanded or initial query, whereas the difference in performance between typical expansions and the unexpanded query was very small in most cases. This suggests that overly aggressive query expansion approaches are ineffective when compared to their less aggressive counterparts. Xu and Croft [16] extended the concept of QE by using the local documents retrieved by the initial query alongside global analyses of the corpus to discover word relationships. Through the process of local feedback, Xu and Croft were able to use the top-ranked local documents to reinforce and alter the probabilities of word relationships to improve the outcomes of future queries.

The need for automated QE has increased due to the explosive increase in the volume of data while the increase in the number of user query terms has remained low [6]. There are two main steps for query expansion. The first step is to find the relationships between queries and words and to select the top related words to expand the query. The second step is to apply the expanded queries for document retrieval, ranking, and computing the final ranking relevance scores [8]. Currently, there are open source tools that can be utilized as a source of related words, commonly used tools include WordNet [17] and Datamuse [13]. This study utilizes the Datamuse API because it embeds WordNet and provides a wide range of features including autocomplete on text input fields, ranking of search relevancy, assistance in writing apps, and word games.

*2) Clustering Search Results*

User queries that express broad search intent often retrieves intractably large result sets. Clustering search results is a technique for organizing search results to achieve coherence, distinctiveness, and clarity. This means that each cluster will represent a distinct and coherent subset of possible search intents and the meaning of each cluster should be clear to the information seeker or user [6]. Search result clustering involves embedding data into vectors and then computing a geometric function on them such as cosine similarity to measure their resemblance [18].

Vector clustering algorithms have existed for a long period of time and are extremely valuable in data science applications. However, in the context of a QE task, there are two factors to consider when choosing/creating a clustering technique. First, the runtime must be considered, as users expect almost instantaneous search results from the Web or other search systems. Secondly, the algorithm must retain reasonable speed when dealing with high dimensional vectors. K-Means, one of the simplest vector clustering techniques, involves choosing a group count (number of centroids) to use beforehand and performing several iterations to move centroids about in space, optimizing the distance between centroids and the respective group data [19]. This is an effective technique for smaller datasets and has relatively low runtime requiring few iterations in most cases. However, density-based methods such as K-Means have limitations when handling high-dimensional data, as the feature space is usually sparse, making it difficult to distinguish high-density from low-density regions [20].

An algorithm well suited for high dimensional datasets is the mean shift technique [21, 22], which iteratively moves a centroid towards an area of higher point density. The base implementation cannot be used for high dimensionality datasets, however, Georgescu et al. adapted the initial algorithm for high dimensions using multiple partitions to approximate the desired cluster [23]. The difficulty with this method in terms of QE, stems from the requirement of providing a radius value for the centroid. Since the number of synonyms/related words for a given cluster is unknown at the beginning of the QE, this hyper-parameter cannot be provided. These constraints mean that pre-existing algorithms for clustering cannot successfully handle the given QE task. Thus, we developed a new algorithm as detailed in Section III.

*3) Query Optimization*

The precision and recall rates of the expanded queries can be improved through several methods. These methods were surveyed by Azad & Deepak and include techniques such as Boolean operators, XML, disambiguation, collection of top terms, and concepts [5]. We applied the Boolean improvement in our work.

*4) Relevance Feedback*

Relevance feedback takes the initial results returned from a given query and applies user feedback about whether or not these results are relevant, to formulate and execute a new query [6]. As such the users are regarded as the ultimate judges of query expansion and result quality [2].

*B. Data Transformation*

Data transformation in the context of this study is the process of representing, preprocessing, and transforming the raw data that will be used for expanding the user query into features, which can be more effectively processed by the subsequent steps [6]. Depending on the nature of the input data source, many data to feature transformation methods exist. A simple way to accomplish this is to use a technique known as One-Hot-Spot representation; a vector with size equal to the vocabulary size, where the word index is denoted with a 1 while the remaining values are 0's [24]. Although this is a valid word representation technique, it has several drawbacks. First, as the vocabulary size increases, so does the size of each word vector, leading to the curse of dimensionality [25]. Secondly and most importantly, this representation ignores any existing word similarities, favouring to simply represent different words as being orthogonal in the vector space [26].

As an alternative, Bengio et al. [25] suggested a neural probabilistic language model where words can have multiple degrees of similarity, which implies that two words having similar meanings would have similar vector representations. This type of representation otherwise known as a continuous vector representation, has significant advantages over the one-hot-spot technique, due to its fixed vector size and ability to model similarities between words numerically. After the initial paper was published discussing this concept, a team at Google successfully implemented a tool called Word2Vec (W2V), which contrasted different models for estimating the continuous vector representations of words and provided a C++ implementation for training and reading these types of models [27]. W2V was an invaluable tool throughout this research as it provided a way for individual words to be represented as vectors.

III. COMPUTATIONAL MODELS USED IN XU

This section explains our proposed automated query expansion technique. We first describe the approach to discover relevant words with respect to the query terms. Then we explain the approach to further optimize the selection of these relevant words using high dimensional clustering and Boolean query formulation technique.

*A. Discover Relevant Words*

The most straightforward approach is to query either WordNet [17], a large lexical database of English cognitive synonyms (synsets), or Datamuse API [13], a word-finding query engine, to retrieve the top suggestions and append them to the initial query using OR logic. We considered the Datamuse API as the best option because it taps from many open data sources including WordNet, W2V, and an online dictionary. For instance, given a query term "Prosecution", we compared the top 8 suggested terms from both WordNet and Datamuse as shown in Table I.

TABLE I. WORDNET VS DATAMUSE QUERY SUGGESTIONS

| Suggestions for "Prosecution" | |
|---|---|
| *Wordnet* | *Datamuse* |
| Legal Action | Pursuance |
| Action | Prosecutors |
| Action at Law | Prosecutor |
| Collection | Prosecuting |
| Aggregation | Retrial |
| Pursuance | Trial |
| Continuance | Criminal |
| Continuation | Prosecuted |

The Datamuse expansion seems better than that of WordNet which contains unrelated words such as "Continuation". Using the OR logic gate, the query term 'prosecution' would be expanded to 'pursuance OR prosecutors OR prosecutor OR prosecuting OR retrial OR trial OR criminal OR prosecuted'. This expansion, although better, would yield extremely noisy results, containing almost any article related to crime, criminal activity, or trials, therefore returning much-unwanted information.

Although the technique had a low runtime cost and was able to handle a variety of difficult queries without failure, the method yielded a significant number of extraneous results, which deviated quickly from the users' initial search query. Also composing the suggested words using only OR logic gate did not fit within the constraints of the QE problem which include handling complex queries with a good response time and noise ceiling. Therefore, additional research was needed to further refine the QE result to be more closely related to that of the initial query. The results also suggested that instead of using the simple OR logic to expand the initial query, a more complex logical combination with AND or NOT may increase the accuracy of the QE tool.

*B. Optimize Selection of Words*

After evaluating the initial approach, it was clear that further refinement was needed to reduce the number of extraneous results generated by the QE. The challenge was to correctly orient similar words into groups for creating the QE string. We first transformed the raw data that would be used for expanding the user query into features represented as vectors to enable effective processing and clustering in the subsequent steps.

*1) W2V Representation*

We used the W2V tool to generate a W2V model containing continuous vector representations of 2,562,529 words from every article on the Wikipedia site. This trained W2V model served as the primary data source in this research and was used to convert words to feature vector representations in 200-dimensional Euclidian space[1], which allowed a clustering algorithm to be effective in finding groups of synonyms. For each user query, words from the query were mapped to vectors using the W2V model. A simple approach to finding synonyms of these words was to compare the vectors of the initial query with every other vector in the W2V model and select the top results. However, this technique had a significant runtime cost, due to the expensive IO operations as well as over 2.5 million comparisons of high dimensional vectors. We addressed this challenge using the Datamuse API to generate a maximum of 100 similar/relevant words for the user queries.

After retrieving the set of 100 suggested words from Datamuse API, we generated the equivalent set of vectors for these words using the W2V model. Next, we compared the two sets of word-vectors, one generated from the words in the user query and the other generated from the suggested 100 words and selected the top n results. This optimization drastically improved the runtime of this step in the QE. An additional benefit of using the Datamuse API was that it supported both singular and compound word queries to a high degree of accuracy, through its extensive use of online resources as its data source.

To further numerically justify the choice between WordNet and Datamuse, we compared the accuracy of these two systems in terms of a similarity Score computed using in Eq. 1. As explained above for each of these systems and a given user query word, we generated a set of n suggested word vectors X and a query word vector I using the trained W2V model. Next, we computed the sum of similarity score for each system by adding the absolute difference of each of the suggested vectors to the query vector as shown in Eq. 1.

$$S(X, I) = \frac{\sum_{W \in X} |W - I|}{n} \qquad (1)$$

The above process was applied for WordNet and Datamuse for three sample user queries and the results are shown in Table II. A lower similarity score indicates a smaller difference between the query and the suggested words, and hence higher similarity and shorter overall distance among the suggested words in the Euclidian space.

TABLE II. NUMERICAL COMPARISON OF WORDNET AND DATAMUSE

| Suggestions for "Prosecution" | | | |
|---|---|---|---|
| *Example* | *WordNet Score* $(S_w)$ | *Datamuse Score* $(S_d)$ | $S_w - S_d$ |
| *Prosecution* | 668 | 575 | 93 |
| *Compute* | 579 | 442 | 137 |
| *Medium* | 698 | 655 | 43 |

On average, Datamuse API gave suggestions which were 91 units closer to the initial search query than WordNet, which was an expected result as Datamuse uses WordNet as a data source alongside many others. From these results, we validated that Datamuse API was the optimal choice for building the QE tool as it would help to significantly reduce the runtime of the QE procedure while simultaneously validating the results of the W2V model.

---

[1] 200 is arbitrary, chosen to reduce required training time and file sizes.

*2) High Dimensional Clustering.*

We used n=25 in the subsequent steps, a set of top n suggested similar words from Datamuse API, which can be changed as necessary. Therefore, the search space was minimized to the top n similar words from Datamuse API in addition to the words from the initial query, and the vector representations for those words could be easily retrieved from the W2V model.

Next the challenge was to find the optimal group orientation. Existing algorithms such as k-means or mean shift have drawbacks that prevent them from being effective in this task, which implies that a new clustering algorithm must be developed. One solution to this problem could be using graph theory: creating a fully interconnected graph, where a node represents a word, and each edge represents the distance between two words. Using this approach, a community detection algorithm, would theoretically be able to compute the optimal groups of words. However, as shown by Lancichinetti et al., these algorithms have a very high runtime cost, especially as the number of nodes increases in the graph [28]. Therefore, we applied a simpler solution by splitting the main problem into two smaller problems. The first problem was to find a numerical method to assess the proximity of vectors within a cluster of any number of vectors. When two vectors are similar in a Euclidian space, their corresponding words have similar meanings, which is a relationship ensured by the W2V trained model. Therefore, the score of a cluster of vectors should reflect how close they are to each other in a vector space.

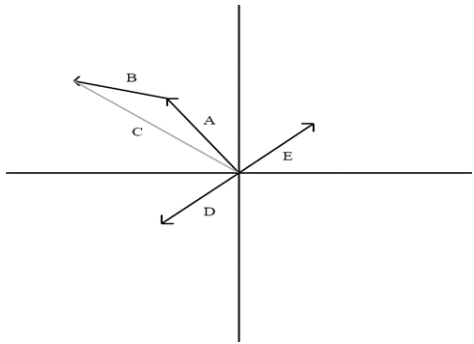

Fig. 1. An illustration of multiple vectors used as an aid for explaining the mathematical basis behind the cluster scoring formula.

As shown in Fig. 1, for the five vectors labelled A through E in a 2D space, C represents the vector sum of A and B, and D and E cancel out the effect of each other since they are mathematically opposite. Ideally, the cluster scoring function would assign a score of 0 to a set of vectors containing D and E since they demonstrate the worst possible grouping of 2D vectors. On the other hand, similar vectors should generate a higher cluster score. C is the sum of A and B in Fig. 1.

Based on these facts, the following formula was developed to compute a cluster score to indicate the quality or orientation of similar vectors within a cluster. As shown in Eq. 2, the score S of a cluster C, containing n vectors, is given by the magnitude of the sum of each vector in the cluster, divided by the sum of the magnitude of each vector. Essentially, vectors which are closely clustered will add to produce a magnitude close to the total sum of the magnitude of each vector. Vectors which are poorly clustered will cancel each other out after addition, therefore reducing the value of the numerator in Eq. 2.

$$S(C) = \frac{|\sum_{i=1}^{n} \overline{c_i}|}{\sum_{i=1}^{n} |\overline{c_i}|} \qquad (2)$$

This function fulfills the requirement of yielding a value of 0 to a vector set of D and E as shown in Fig. 1.

Now that a cluster, which is defined as a group of words, can be scored numerically, all that remains is to develop some method to find the optimal orientation of a group of clusters. Since the number of clusters m, can be provided as a hyper-parameter, the numerical score of a group of clusters $S_g$ can be computed simply by adding the individual scores $S_i$ of each cluster as shown in Eq. 3. Accordingly, a group of clusters that has the optimal orientation is the one with the highest overall score $S_g$. Eq. 2 and 3 accomplish the two requirements for a QE clustering technique discussed earlier by enabling it to effectively cluster high dimensional vectors while maintaining a reasonable runtime speed.

$$S_g = \sum_{i=1}^{m} S_m \qquad (3)$$

*3) Boolean Query Formulation*

Clustering divides the word vectors into a few groups where vectors in each group have similar meaning. In the next step, we use Boolean operators to formulate queries with a view to enhancing both recall and precision rates. First, words in a cluster having similar meanings are ideally joined with OR, because an article should pass that filter if it contains either of the similar words. Second, these groups of OR-ed words or each cluster of words can be combined with other logic such as AND or NOT to narrow the search results and add multiple layers of filtering before a document is returned. We illustrate the Boolean query formulation using the same example query term "Prosecution" as given below.

Prosecution: ('indictment' OR 'attorneys' OR 'allegation' OR 'counsel' OR 'incrimination' OR 'complainant' OR 'prosecuting' OR 'prosecute' OR 'trial' OR 'impeachment' OR 'judge' OR 'proceedings') AND ('prosecutorial' OR 'evidentiary' OR 'pursuance' OR 'retrial' OR 'arraignment' OR 'prosecuted' OR 'indictments' OR 'prosecutors' OR 'conviction' OR 'prosecutor' OR 'charges' OR 'criminal' OR 'punishment' OR 'prosecution'). A generic template depicting this idea is shown in Fig. 2.

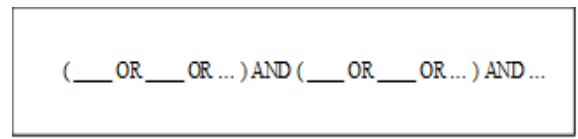

Fig. 2. A blank template outlining a Boolean logic statement that would provide multiple layers of filtering before a search result is returned. Similar words are grouped with OR gates, and multiple groups are combined with restrictive gates.

The overall architecture of Xu is shown in Fig. 3 which performs QE in the following steps.
- The original user queries are preprocessed and then utilized to generate candidate expansion terms obtained using the Datamuse API.
- Selection of candidate terms is further optimized as follows.
- Large document collections such as Wikipedia is used to create a W2V model to generate word embedding for the above terms.
  - To reduce the level of noisy or unrelated terms, a high dimensional clustering method is used to group similar word vectors into clusters. The vector sum of words is used to compute a cluster score where higher score indicates a better cluster containing more similar words. The best group of clusters are selected based on a group similarity score which is the sum of the scores of its clusters.
  - To further reduce the noise, a simple Boolean query formulation and optimization approach is applied to compose the words from the best group of clusters having the highest score.

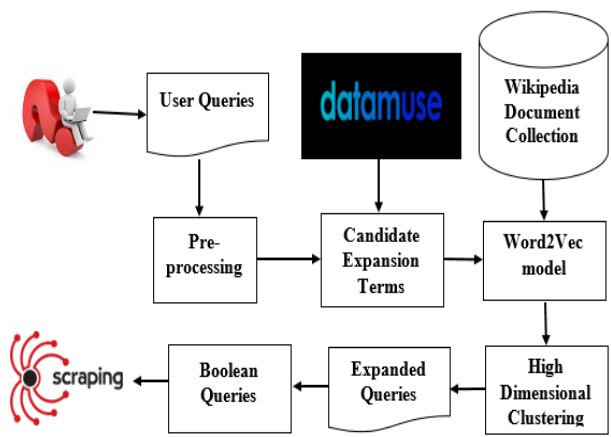

Fig. 3. The overall architecture of Xu.

In the next section, we illustrate the implementation of our QE tool Xu, which applies the steps explained in this section.

## IV. IMPLEMENTATION

Many open source libraries were used in implementing Xu as described in this section with footnote references to the respective online links and repositories. Xu enables query expansion and optimization with Boolean enhancement functionality for both a single query and multiple queries listed in a CSV document as provided to us by our industry collaborators. The CSV file contained manually formulated query expansion given specific query terms. The goal was to automate the query expansion process for the given query terms. Xu includes the W2V tool for training the required W2V word vector model. Xu has been built and tested successfully on Ubuntu 16.04 & Ubuntu 18.04 systems.

### A. Open Source Libraries

The Xu command line tool for automated QE was implemented using the C++ programming language. It includes a C++ implementation of the Word2Vec[2] library. This Word2Vec library is based on the original paper from Google with enhancements to parallelize the training of the model using a commonly used high-performance API known as OpenMP [29]. Xu is integrated with the Word2Vec implementation through a command line interface, which allows users to train a W2V model by passing in a corpus of text. The text corpus must have been pre-processed to contain no punctuations apart from periods. An open source C++ JSON parser[3] is also used in Xu in conjunction with the libcurl[4] library to efficiently process the JSON document returned by the Datamuse API as a response to a get request.

### B. Using Xu

Xu is self-contained. It requires a word vector model file to be passed in as a parameter, which it initially loads in the memory for quick word matching and vector transformations. Once the model file is loaded, given the user query terms, Xu retrieves a 200-dimensional vector transformation, I, of the original user query terms. Next, Xu proceeds to query Datamuse API for the matching word list. It saves the returned JSON document from Datamuse and parses it to retrieve the array of potential matching words. Once a list of top N words is chosen from Datamuse API to be used in query expansion, Xu uses the trained W2V model to retrieve the corresponding set X of 200-dimensional transformed word vectors W. A similarity score S(W, I) is computed next for each suggested word W using Eq. 4 where W $\epsilon$ X. We used N=50 in our study[5].

$$S(W, I) = |W - I| \qquad (4)$$

Top n word vectors W with the highest score are used to redefine X and are used in the QE. We used a set of top 25 vectors (n=25), which are then used in the high dimensional clustering algorithm as described in Section III. The clustering algorithm groups these vectors into M groups, where M is provided as a hyper-parameter. This allows Xu to choose which groups of words should be combined using OR and AND logic. The generated Boolean queries can either be saved to a file or be utilized, for example, in a search application or in a web or document scraping system as shown in Fig. 3, to retrieve relevant documents or articles from a collection.

With the current implementation, initial queries which are very long or Out of Vocabulary (OOV) may not be correctly expanded. Next, we present experimental results on evaluation of the performance of Xu and several optimizations strategies that were carried out to improve the tool.

---

[2] https://github.com/jdeng/word2vec
[3] https://github.com/open-source-parsers/jsoncpp
[4] https://curl.haxx.se/libcurl/
[5] A value of 50 was chosen arbitrarily.

## V. IMPLEMENTATION

This section focuses on the evaluation of the runtime speed of Xu and the clustering algorithm using a list of query terms provided by our industry collaborator. The runtime of Xu would be important when using it as a search engine. The QE generation time also adds an overhead due to need for filtering each document in the corpus. The query terms were provided as a string of words in a CSV file and were manually expanded to form new query strings to retrieve relevant documents from the web to send to the customers. The goal of this exercise is to use Xu to expand a set of given user query terms, and thereby, create an expanded Boolean query with these query terms.

### A. Dataset

A media monitoring company provided us with a CSV file containing 373 unique user queries, which we used to validate Xu. A total of 241 of the queries had corresponding expanded query that were used by the company, these helped us validate the accuracy of Xu.

### B. Hardware and Network Requirements

The list of queries to expand was distributed over 12 logical processors on an Intel Core i7-8750H CPI, with a base clock of 2.20 GHz, using the OpenMP C++ library. At the time of experimentation, the internet connection speed was 260.45 mbps download, 284.55 mbps upload, and 19 ms ping.

### C. Runtime Speed

First, the runtime of the Datamuse API query was evaluated. For each of the 373 queries in the CSV document, the time to successfully send the get request to Datamuse, as well as parse the returned JSON document was recorded and averaged. This experiment was completed three times to ensure consistency of the results. On average, it took 247.6 ms (milliseconds) per word to complete this task, with a standard deviation of 50.7 ms.

The second experiment involved altering the number of suggestions (word vectors) that were processed by the clustering algorithm, to detect changes in query expansion runtime speed because of a change in the number of words. For each experiment, three groups of vectors were generated, and the clustering algorithm performed 10,000 iterations. In total, three trials were conducted for each data point.

Fig. 4 displays a line graph showing the runtime speed per query because of a change of word count. The graph displays an exponential slope, meaning that as the number of vectors to be processed by the clustering algorithm increases, the runtime of the algorithm increases exponentially. This was anticipated as each iteration requires more computation. However, as stated by Voorhees [15], aggressively expanded queries perform worse than expansions with fewer additions [11]. Therefore, increasing the number of words is not an ideal optimization strategy as increasing the word count would lead to less accurate results.

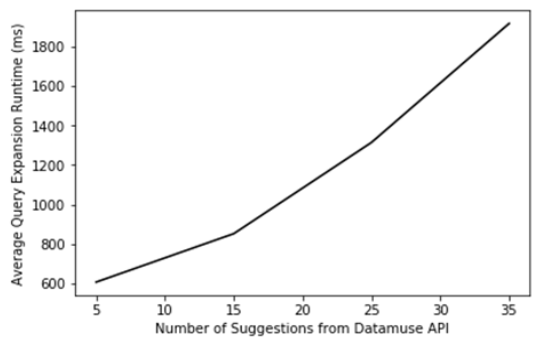

Fig. 4. A line graph depicting the relationship between the number of suggestions retrieved from Datamuse API, and the average query expansion runtime speed.

Last, we studied the effect of increasing the iteration of the clustering algorithm on the runtime of query expansion. For each experiment, three groups of vectors were generated, using 25 suggestions from Datamuse API, along with the initial query. In total, three trials were conducted for each data point to ensure consistency in the results. Fig. 5 shows a graph depicting this relationship.

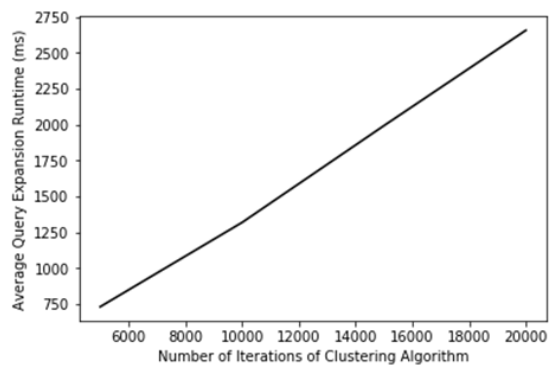

Fig. 5. A chart displaying the relationship between the number of iterations of the vector clustering algorithm, and its effect on the query expansion runtime.

As shown in Fig. 5, the resultant slope is linear, meaning that the number of iterations changes the query expansion runtime at a constant rate. This allows the user a significant amount of flexibility to choose a shorter runtime versus potentially a more accurate grouping of the words.

### D. Quality of Expanded Query Results

This section discusses the evaluation of the quality of the results returned by automatically expanded queries generated by Xu against those defined by a human expert. In practice, the quality of search results is mostly judged by humans, which is prone to subjective bias. Therefore, rather than looking at an individual query expansion and evaluating its effectiveness on arbitrary metrics, we decided to assess the automated expansions on a relative scale, compared to human-generated expansions. A numerical score is given to each automated query expansion, when there is an accompanying human (base) QE, which is assumed to be optimal[6]. For this experiment, three different query expansions were tested: a) human generated or base, b) using only relevant words from Datamuse without further optimization, and c) optimized query expansions generated by Xu using additional clustering and query formulation techniques.

---

[6] As it is being used in a propriety news monitoring system.

To effectively evaluate a wide range of query expansions, which covered a wide variety of subject areas, a dataset was needed that covered a variety of topics, and that represented a use case of a search engine. For this, we used a publicly available dataset, called "All the News"[7], which contains 146,032 news articles from 15 different modern publishers such as the New York Times, CNN, and the Washington Post. The articles pertain to a multitude of different topics, such as politics, business, technology and the environment. For every example query which had an accompanying human generated QE, the results from the three QE techniques were compared by running each QE against each news article. For each of the techniques, a scoring vector of size equal to the number of news articles was generated, containing Boolean values of 1s or 0s for each article. For the human generated QE, the position for an article in the scoring vector contained 1 if the article was selected and 0 otherwise. For the other two QE techniques, the position for an article in the scoring vector contained 1 if the article was selected by both the human (base) QE and the other QE techniques (Datamuse and Xu) being compared, and 0 otherwise, the orders of the articles in all scoring vectors being the same. A true positive score is computed using Eq. 5 where BO is the count of Boolean 1s for other QE techniques in the scoring vector and BH is the same for the human generated QE.

$$TP = \frac{\sum_{i=1}^{c} B_O}{\sum_{i=1}^{c} B_H} * 100 \qquad (5)$$

Fig. 6 shows a plot displaying the distribution of scores from these three approaches.

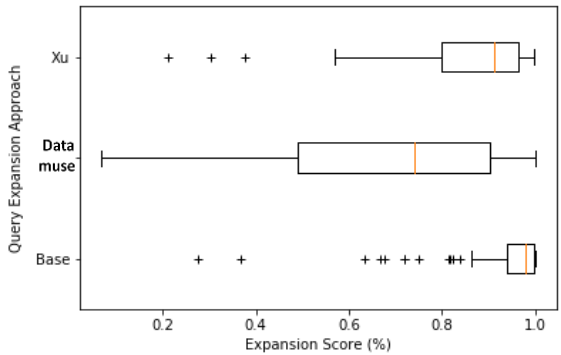

Fig. 6. A set of boxplots displaying the distribution of scores from 132 queries, meant to compare the base (human) query effectiveness, versus that of the automated initial approach, as well using Xu.

As shown in Fig. 6, the expansions from Xu performed significantly better on average than those of the Datamuse approach. This comparison is only relative to the human (base) QEs. We also evaluated the results in terms of the number of negative events wrongly categorized as positive (false positives). In the context of this research, false positive means a document was selected when it should not have been based on human (base) QE results. The percentage of false positives FP was calculated using Eq. 6 where FPO is the count of false positives selected by the other QE techniques and FH is the count of documents not selected by human (base) QE.

$$FP = \frac{\sum_{i=1}^{c} FP_O}{\sum_{i=1}^{c} F_H} * 100 \qquad (6)$$

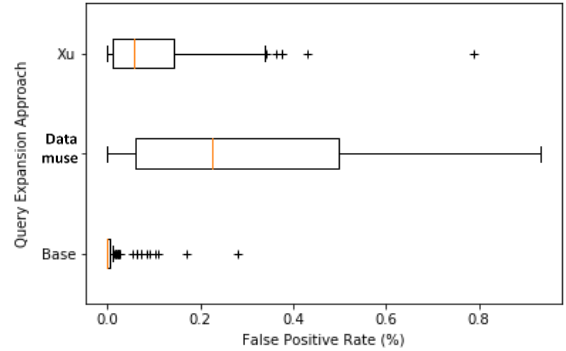

Fig. 7. A set of box plots showing the distribution of false positive rates from 132 queries, used to show the difference between the varying approaches.

Fig. 7 shows the false positive rates of the three approaches. The figure is presented as a box a plot showing the quartile values, median, and the presence of outliers. The lower the false positive rate, the better the approach. Fig. 7 validates the conclusions made in Section 3 that many extraneous results are returned by the simple Datamuse QE approach. Optimization using the Boolean query formulation by the Xu command line tool reduces the false positive rate of the query expansions. This conclusion is similar for false negative rates. The accuracies of the three QE approaches (Base, Datamuse, and Xu) were computed with reference to the true positive (TP), true negative (TN), false positive (FP), and false negative (FN) values using equation 7 described by Olson & Delen [30].

$$accuracy = (TP + TN)/(TP + TN + FP + FN) \qquad (7)$$

The runtime speed for processing the 132 queries for the different QE approaches were recorded during the evaluation. The average accuracies of the three different QE approaches as well as the averaged time elapsed are shown in Table III). The average accuracy for the Base or human generated QE in Table III is 1 or 100% because it was considered as the standard for comparison. Our proposed QE tool, Xu, proved to be superior to the Datamuse approach with an average accuracy of 0.88 or 88% as against the Datamuse QE approach which realized an accuracy of 0.70 or 70%.

TABLE III.    EVALUATION RESULT USING THE 132 QUERIES

| Metric | Base | Datamuse QE | Xu QE |
|---|---|---|---|
| *Average accuracy* | 1 | 0.70 | 0.88 |

---

[7] https://www.kaggle.com/snapcrack/all-the-news

| | | | |
|---|---|---|---|
| *Average time elapsed (s)* | 116.3 | 108.1 | 170.5 |

Finally, the average time elapsed indicate that as the complexity of the query expansion increases, so does the runtime (Xu having the highest average run time of 170.5 seconds per query). This must be considered when generating QE, as smaller expansions can be processed faster by a search engine.

Queries in information retrieval are the primary ways in which information seekers communicate with search and retrieval systems. Search engines support two major functions, indexing and query processing. The indexing process builds the structures that enable searching while the query process utilizes those structures and users' query to produce a ranked list of results. One of the main goals of information retrieval systems is to satisfy user information needs by instantly returning relevant results to a given query. However, user queries are usually too short to express appropriately what the user is looking for, or the query expressions are not well formulated, or the context is not correctly presented in the query. These are some of the reasons why search results often do not meet user expectations.

## VI. CONCLUSIONS

In this paper, we present an automated query extension tool called Xu, which we designed and implemented as an open-source self-contained command line tool for media analytics data acquisition. To achieve query expansion and optimization, Xu utilizes Datamuse, an open source API, for finding semantically matching and relevant words, and a function for ranking and selecting top k-word suggestions. A Wikipedia-trained Word2Vec model is used by Xu to generate vector representations of words to be used in a high dimensional clustering algorithm that generates clusters of similar word vectors. Finally, Xu applies a model for formulating expanded queries using Boolean operators to compose the groups of word vectors into a query string of words. By considering human generated QE to be optimum, we evaluated Xu by comparing the results returned by expanded queries created by humans, by Datamuse, and by Xu. The dataset containing human-expanded queries was provided to us by a media analytics company. The expanded queries were run against an open source dataset containing news articles. The evaluation results and comparison between Xu and Datamuse QEs indicate that Xu perform better than Datamuse. Therefore, instead of manually generating the QE, Xu can be used for efficient IR.

The assumption that human generated QE is optimum is impractical as it is prone to human error and bias. However, validation of the accuracy of IR is inherently a difficult problem due to the subjective bias. As future work, we will extend Xu to a full-fledged search engine that can be embedded in other applications such as assistive systems, decision support systems, chatbots or voice-enabled interactive cognitive systems. A major limitation that must be resolved is the high runtime of the QE generation. One suggestion is to incorporate a caching mechanism for keeping precomputed results in memory to reduce the computational cost of frequent read/write operations. The current implementation of Xu is limited to the use of only AND logic gates as restrictive gates, which can hinder the accuracy. The use of other restrictive gates such as NEAR and NOT should be explored. Additionally, experiments utilizing user feedback and comparing the effectiveness of changes in dimensionality, iteration count and word count on the accuracy of query expansions would be valuable.

ACKNOWLEDGMENT

We like to express special thanks to Southern Ontario Smart Computing for Innovation Platform (SOSCIP) and IBM Canada for supporting this research. We are also grateful to Queen's University Centre for Advanced Computing (CAC) for providing access to computing resources to train the W2V vector model.